# An evaluation of GPT models for phenotype concept recognition

Tudor Groza, PhD[1,2,3,4,*], Harry Caufield, PhD[5], Dylan Gration, BSc[6], Gareth Baynam, PhD, MD[1,2,6,7], Melissa A Haendel, PhD[8], Peter N Robinson, PhD, MD[9,10], Christopher J Mungall, PhD[5], Justin T Reese, PhD[5]

1. Rare Care Centre, Perth Children's Hospital, 15 Hospital Avenue, Nedlands WA 6009, Australia
2. Telethon Kids Institute, 15 Hospital Avenue, Nedlands WA 6009, Australia
3. School of Electrical Engineering, Computing and Mathematical Sciences, Curtin University, Kent St, Bentley WA 6102, Australia
4. SingHealth Duke-NUS Institute of Precision Medicine, 5 Hospital Drive Level 9, Singapore 169609, Singapore
5. Division of Environmental Genomics and Systems Biology, Lawrence Berkeley National Laboratory, Berkeley, CA 94720, USA
6. Western Australian Register of Developmental Anomalies, King Edward Memorial Hospital, 374 Bagot Road, Subiaco WA 6008, Australia
7. Faculty of Health and Medical Sciences, University of Western Australia, 35 Stirling Hwy, Crawley WA 6009, Australia
8. University of Colorado Anschutz Medical Campus, Aurora, CO 80045, USA
9. The Jackson Laboratory for Genomic Medicine, Farmington CT, 06032, USA
10. Institute for Systems Genomics, University of Connecticut, Farmington, CT 06032, USA

* correspondence to: Rare Care Centre, Perth Children's Hospital, 15 Hospital Avenue, Nedlands WA 6009, Australia; tudor.groza@health.wa.gov.au

## ABSTRACT

**Objective:** Clinical deep phenotyping and phenotype annotation play a critical role in both the diagnosis of patients with rare disorders as well as in building computationally-tractable knowledge in the rare disorders field. These processes rely on using ontology concepts, often from the Human Phenotype Ontology, in conjunction with a phenotype concept recognition task (supported usually by machine learning methods) to curate patient profiles or existing scientific literature. With the significant shift in the use of large language models (LLMs) for most NLP tasks, we examine the performance of the latest Generative Pre-trained Transformer (GPT) models underpinning ChatGPT as a foundation for the tasks of clinical phenotyping and phenotype annotation.
**Materials and Methods:** The experimental setup of the study included seven prompts of various levels of specificity, two GPT models (gpt-3.5-turbo and gpt-4.0) and two established gold standard corpora for phenotype recognition, one consisting of publication abstracts and the other clinical observations.
**Results:** Our results show that, with an appropriate setup, these models can achieve state of the art performance. The best run, using few-shot learning, achieved 0.58 macro F1 score on publication abstracts and 0.75 macro F1 score on clinical observations, the former being comparable with the state of the art, while the latter surpassing the current best in class tool.
**Conclusion:** While the results are promising, the non-deterministic nature of the outcomes, the high cost and the lack of concordance between different runs using the same prompt and input make the use of these LLMs challenging for this particular task.

# INTRODUCTION

Over the past decade, clinical deep phenotyping - i.e., the comprehensive documentation of abnormal physical characteristics and traits in a computationally-tractable manner - has evolved into a common procedure for individuals who are either suspected of having or have been diagnosed with a rare disease. Similarly, the development and continuous enrichment of knowledge bases in the rare disease domain has become standard practice. Conceptually, both tasks rely on ontologies that are developed and updated by the medical community. Such ontologies facilitate the description of a patient's unique phenotype, as well as the characterisation of the phenotypic manifestations of gene mutations using ontological terms and concepts. The utility of using ontology-coded knowledge in rare diseases has been showcased repeatedly over the years in data sharing [1-3] and clinical variant prioritization and interpretation [4-6].

The Human Phenotype Ontology (HPO) [7,8] provides the most comprehensive resource for computational deep phenotyping and has become the de facto standard for encoding phenotypes in the rare disease domain, for both disease definitions as well as patient profiles to aid genomic diagnostics. The ontology, maintained by the Monarch Initiative [9], provides a set of more than 16,500 terms describing human phenotypic abnormalities, arranged as a hierarchy with the most specific terms furthest from the root, as depicted in Fig. 1.

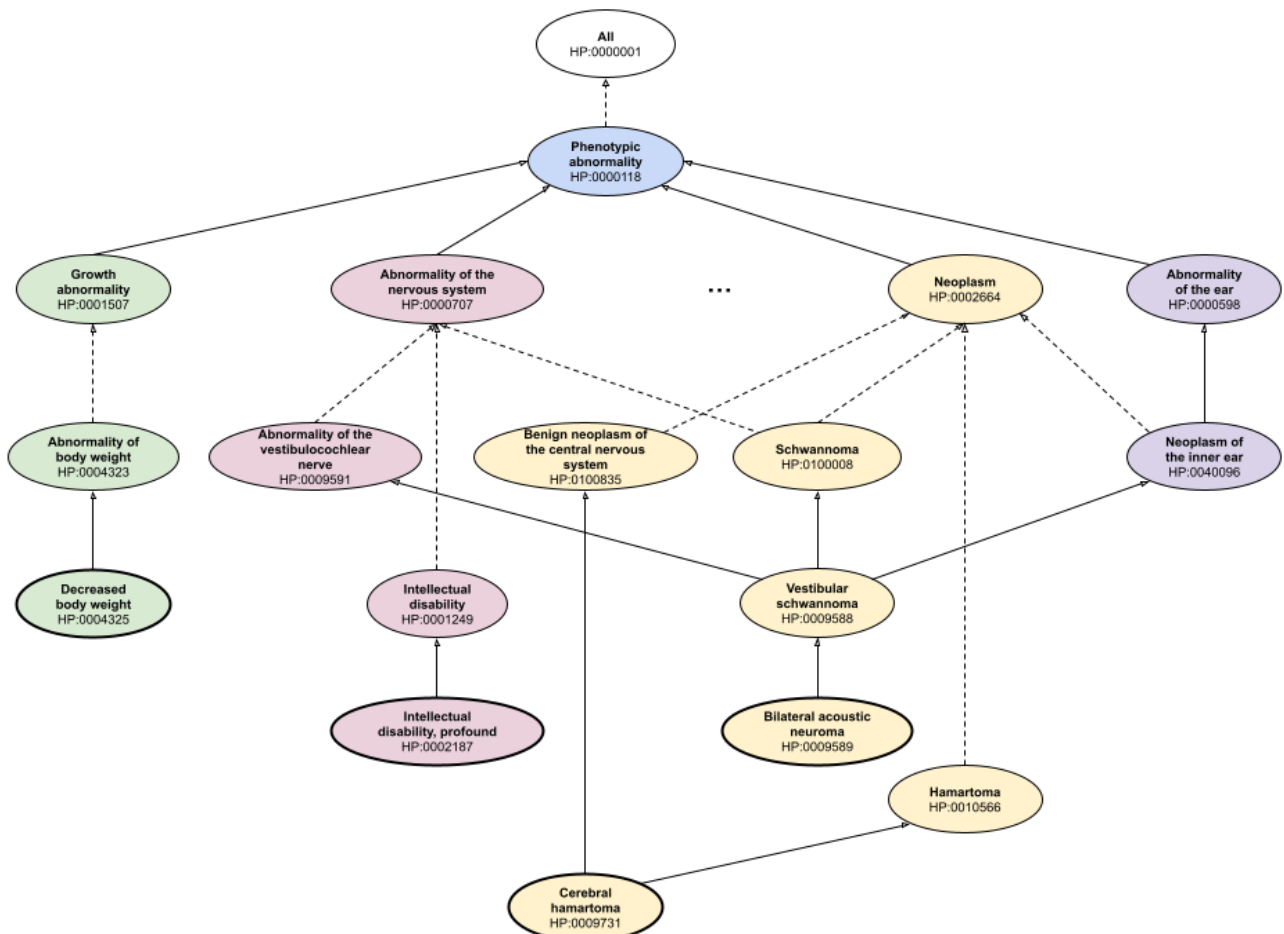

**Fig 1.** Simplified example of Human Phenotype Ontology concepts and their structural arrangement in the hierarchy. Solid lines denote direct parent-child relationships, while dotted lines denote ancestor-descendant relationships.

In addition to underpinning complex diagnostic tasks (e.g., clinical interpretation of an exome/genome) [10] or building care coordination plans, ontology-coded phenotypes often represent also a communication channel between practitioners and patients and, subsequently, between patients and other stakeholders - e.g., education, disability or welfare workers. Moreover, concepts grounded in HPO provide the explainability required to improve the transparency of the decision-making process, which can then support communication and documentation.

Manual curation of phenotype profiles – and manual annotation as a general task – is, however, tedious and has represented the main blocker to wide-spread uptake of computational deep phenotyping on the clinical side and to keeping rare disease knowledge bases up to date. (Semi-)automated methods that rely on natural language processing (NLP) have been introduced to remove this blocker and have gradually become the standard *modus operandi*. Such methods, the latest built using convolutional neural networks [11] or transformer-based architectures [12], are also addressing a variety of challenges associated with phenotype concept recognition (CR) such as ambiguity, use of metaphorical expressions, as well as negation and complex or nested structures.

Lately, the focus has shifted to large language models (LLMs) for most NLP tasks. LLMs - a class of transformer-based models trained on trillions of words of diverse texts [13] - showcased superior capabilities in application domains such as chatbots and text prediction [14]. Their main advantage also stems from having the ability to use few-shot learning to perform specific tasks, without the need for further training or fine-tuning [15], which replaces the "traditional" task-driven training of machine learning models [16]. gpt-3.5 and gpt-4.0 are examples of such LLMs that have witnessed a rapid general user adoption via the ChatGPT application, a chatbot fine-tuned for conversation-based interactions with humans. A user can "prompt" ChatGPT to perform a variety of tasks, with or without the need to provide examples to support them.

In the biomedical domain, several domain-specific models have been published - BioBERT [15], PubMedBERT [17] or BioGPT [18] – and shown to perform well on NLP tasks including relationship extraction (e.g., drug-drug interactions or drug-target interactions) and question answering. The experimental results also included comparisons against GPT-2.0, a predecessor of the current models powering ChatGPT. Lately, several studies have been published on the utility of using GPT models for annotation (in general) [19, 20] and few discussed the efficiency of such models on concept recognition tasks, in particular phenotype concept recognition. Note that ontology-based concept recognition implies a joint task of named entity recognition (i.e., finding entities on interest in a text and their corresponding boundaries) and entity linking (i.e., aligning the entities found in the text to concepts defined in a given ontology). Experiments documenting the accuracy on named entity recognition - with a focus on diseases and chemical entities - were documented by Chen et al. [21], with gpt-4.0 (+ one-shot learning) achieving a performance poorer than a fine-tuned PubmedBERT, yet significantly better that gpt-3.5.

This paper examines the ability of gpt-3.5 and gpt-4.0 to perform phenotype concept recognition using HPO as a background ontology. Three different approaches are used to generate prompts to gain a deeper understanding of the limitations in various scenarios. Specifically, the experimental setup targets direct concept recognition – i.e., named entity recognition followed by an alignment to HPO concepts and few-shot learning.

## MATERIALS AND METHODS

The study uses two gold standard corpora available in the literature for phenotype concept recognition: (i) a corpus of 228 scientific abstracts collected from PubMed, initially annotated and published by Groza et al [22], and subsequently refined by Lobo et al [23] (named HPO-GS from here on); and (ii) the dev component of the corpus made available through Track 3 of BioCreative VIII (454 entries), focusing on extraction and normalization of phenotypes resulting from genetic diseases, based on dysmorphology physical examination [24] (named BIOC-GS from here on). An example of an entry is: "ABDOMEN: Small umbilical hernia. Mild distention. Soft."

All experiments were conducted using these two corpora. HPO-GS covers 2,773 HPO term mentions and a total of 497 unique HPO IDs, with the minimum size of a document being 138 characters, the maximum size 2,417 characters and the average being ~500 characters. BIOC-GS covers 783 HPO term mentions and a total of 358 unique HPO IDs, with the minimum size of an entry being 13 characters, the maximum 225 characters and the average ~56 characters. Note that we chose the dev component of Track 3 because of its similarity in the number of unique HPO IDs and its profile (described below) to HPO-GS. We were unable to download the test component of Track 3 and hence the results reported here are not comparable to the results published by the Track's organisers.

The complexity of the lexical representations of the HPO concepts can be partially assessed based on their length (presented in Fig. 2) and structural placement in the ontology. The latter is depicted in Fig. 3 using the children of the *Phenotypic abnormality* concept as major categories and the values representing the proportion of terms belonging to each category (as also depicted in Fig 1). It can be observed that the large majority of concepts in both corpora (~88%) have low to moderate lexical complexity, with a label length of 4 words or less, and are placed predominantly in the nervous and musculoskeletal system (including here also head and neck and limbs) - i.e., denoting finger, toe, face, arm and leg abnormalities.

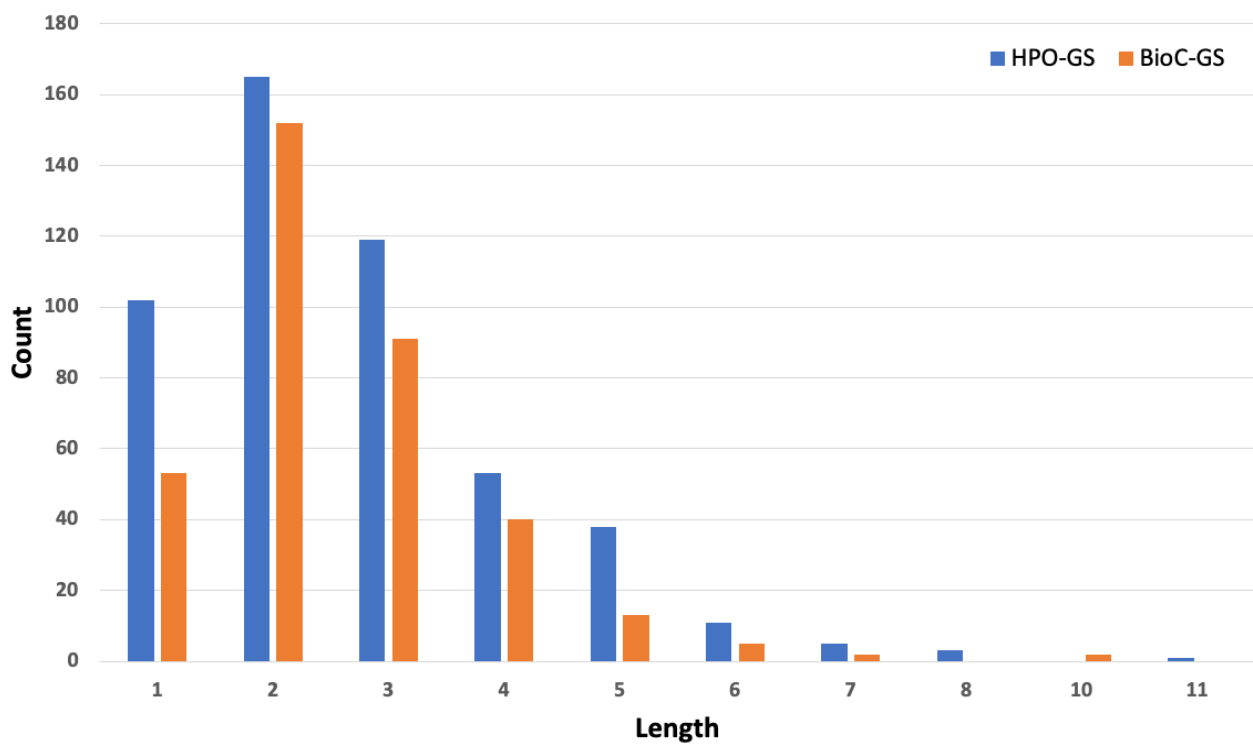

**Fig 2.** Label length distribution of the HPO concepts present in the gold standard corpus

**Prompt generation approaches**

Conversational models such as gpt-3.5-turbo and gpt-4.0 take inputs in the form of prompts. These prompts can include - in addition to a target content - explicit instructions or examples of the desired output. This is sometimes referred to as "prompt engineering". A task, such as concept recognition, can be defined via prompts in various ways, with the behaviour and hence the output of the model being heavily influenced by smallest differences in these definitions. In this study, we used three types of prompts to investigate the models' efficiency to perform phenotype concept recognition. Two remarks are worth noting about our selection strategy:
- We opted for well-known, low-barrier prompts that do not require significant prompt engineering knowledge and skills
- We were aware of the HPO ID hallucinations as a result of aiming for concept recognition – instead of a chain of entity recognition, external entity linking and LLM-based validation, however, as presented later in our experiments, this did not materialise as a real concern.

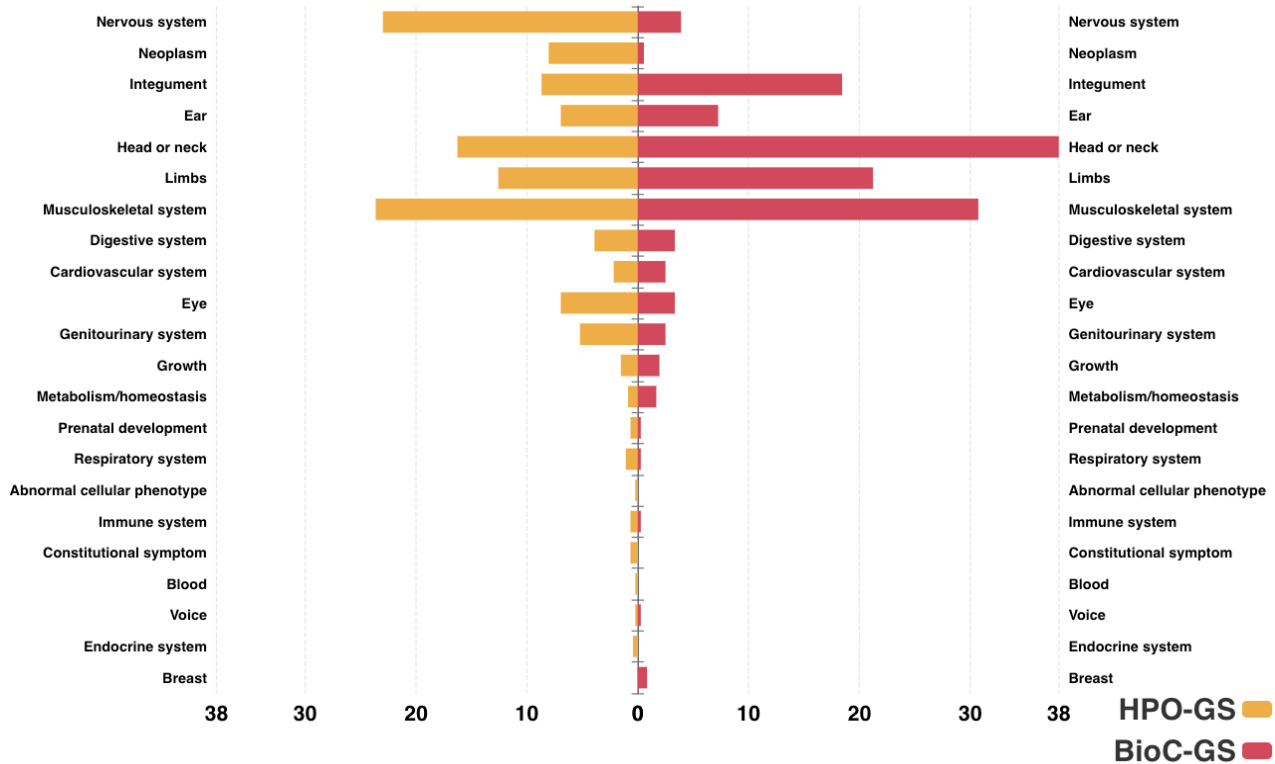

**Fig 3.** Top-level overview of the gold standard corpus using the children of 'Phenotype abnormality' as major categories.

Instructional (directed) phenotype concept recognition. Prompts in this category aimed to capture the impact of the wording used to define 'phenotypes' on the CR task. They are instructional (directed) because the model is asked explicitly to perform a certain task. The four prompts defined in this category are listed below; the key instructions are underlined for easier comprehension.

- Prompt 1: *Analyze the text below delimited by triple backticks and extract phenotypes and clinical abnormalities. Align the phenotypes and clinical abnormalities found to Human Phenotype Ontology IDs. List the results in a JSON format using the following structure.*
- Prompt 2: *Analyze the text below delimited by triple backticks, extract phenotypes and align them to Human Phenotype Ontology IDs. List the results in a JSON format using the following structure. Where you cannot find a direct Human Phenotype Ontology ID, leave the "hpoId" field empty.*
- Prompt 3: *Analyze the text below delimited by triple backticks and extract Human Phenotype Ontology terms. List the HPO IDs together with the start and end offsets.*
- Prompt 4: *You will be provided with a text in triple backticks. The task is to perform automated concept recognition using the Human Phenotype Ontology and extract all Human Phenotype Ontology concepts found in the text. Include the HPO ID of the concepts you find in the result.*

The first three prompts direct the model to 'extract' artefacts from the provided text. Prompt 2 is a variation of Prompt 1 ('phenotypes' vs 'phenotypes and clinical abnormalities'), while Prompt 3 refers directly to HPO terms. Prompt 4 explicitly names the task requested to be performed by the model - i.e., 'automated concept recognition'.

Instructional (directed) named entity recognition followed by instructional (directed) entity alignment. The prompts in the first category target directly concept recognition by requesting HPO IDs. As a task, concept recognition can also be modelled as named entity recognition (used to detect entity boundaries in the text) followed by entity alignment (used to match the candidates extracted from the text to ontology concepts / IDs). Prompt sets 5 and 6 below explicitly perform this two-step process by first asking the model to extract phenotypes, then using this output as input to align the text to HPO IDs. Prompt set 6 is a subset of Prompt set 5 – 'phenotypes' vs 'phenotypes and clinical abnormalities'.

- Prompt set 5:
    - Step 1: *Analyze the text below delimited by triple backticks and <u>extract phenotypes and clinical abnormalities</u>. List them together with the start and end offsets.*
    - Step 2: *You will be provided with text delimited by triple backticks. <u>Align the text below to Human Phenotype Ontology labels</u>. List only the HPO concepts found.*
- Prompt set 6:
    - Step 1: *Analyze the text below delimited by triple backticks and <u>extract phenotypes</u>. List them together with the start and end offsets in the text.*
    - Step 2: *You will be provided with text delimited by triple backticks. <u>Align the text below to Human Phenotype Ontology labels</u>. List only the HPO concepts found.*

<u>Few-shot learning using a subset of HPO.</u> The final category (prompt set 7) attempts to aid the model by providing examples of the concepts targeted for extraction. The prompt used a standard two-part template, as below:
- Part 1: Examples: The Human Phenotype Ontology defines phenotype concepts using the following label – HPO ID associations: *Hypospadias* // HP:0000047 …
- Part 2: Task: Using the list above, find Human Phenotype Ontology concepts in the following text and return their associated IDs for every appearance in the text.

Part 1 was completed by adding the pairs of label - HPO ID for all HPO concepts present in the gold standard corpus. Attempts were made to include the entire ontology, or to include the labels and all synonyms for the desired HPO concepts, however they failed due to the model limitations on the size of the input content. We do, however, demonstrate the impact of using various sets of concepts to underpin the few-shot learning task. A complete example of prompt 7 is provided in Appendix 1 in the Supplementary material.

**Experimental setup**

Experiments were conducted by calling the GPT models using the OpenAI API (https://platform.openai.com/docs/api-reference). Each call used one of the seven prompts discussed above and the text corresponding to each abstract or examination entry, individually, as user input. The results were stored individually and HPO concepts were extracted and associated with the PMID / entry ID corresponding to the text used as input. The code used to annotate the corpora and perform the evaluation is available at: https://github.com/tudorgroza/code-for-papers.

The standard evaluation procedure for concept recognition covers two aspects: (i) boundary detection - i.e., a correct alignment of the boundaries of the concepts in text, usually by matching the offsets of the corresponding text span to the offsets found by the system being evaluated; and (ii) concept mapping - i.e., a correct matching of the ID of the concept against that provided by the gold standard. The boundary detection step proved to be challenging to evaluate accurately with the results produced by the OpenAI GPT models – an aspect documented also by Chen et al. [19]. Consequently, given our focus on understanding the utility of these models to support manual phenotype annotation / curation, we relaxed the evaluation procedure to include only the second step - i.e., concept mapping. A correct match was, therefore, counted if the HPO ID present in the gold standard was found at least once by the LLM.

The evaluation metrics used in this experiment are the standard for the task: precision, recall and F1. These were computed at both macro and micro levels. The macro level defines a true positive when a desired HPO ID is found at least once by the LLM, while the micro level keeps track of all encounters of the HPO ID in a particular abstract and defines a true positive when each individual encounter is found by the LLM.

## RESULTS

### Experimental results

| GPT version | Prompt | Precision | Recall | F1 | Precision | Recall | F1 |
|---|---|---|---|---|---|---|---|
| | | **Macro-level** | | | **Micro-level** | | |
| 3.5 | 1 | 0.45 | 0.21 | 0.29 | 0.39 | 0.14 | 0.20 |
| | 2 | **0.51** | 0.12 | 0.19 | **0.46** | 0.08 | 0.13 |
| | 3 | 0.12 | 0.25 | 0.16 | 0.05 | 0.15 | 0.07 |
| | 4 | 0.12 | 0.28 | 0.16 | 0.07 | 0.17 | 0.10 |
| | 5 | 0.14 | 0.09 | 0.11 | 0.14 | 0.06 | 0.08 |
| | 6 | 0.3 | 0.13 | 0.18 | 0.29 | 0.08 | 0.12 |
| | 7 | 0.41 | **0.41** | **0.41** | 0.28 | **0.25** | **0.26** |
| 4 | 1 | 0.41 | 0.34 | 0.37 | 0.36 | 0.21 | 0.26 |
| | 2 | 0.41 | 0.34 | 0.37 | 0.36 | 0.21 | 0.26 |
| | 3 | 0.37 | 0.31 | 0.33 | 0.34 | 0.19 | 0.24 |
| | 4 | 0.34 | 0.38 | 0.35 | 0.32 | 0.23 | **0.27** |
| | 5 | 0.31 | 0.22 | 0.25 | 0.26 | 0.13 | 0.17 |
| | 6 | 0.35 | 0.17 | 0.22 | 0.29 | 0.10 | 0.15 |
| | 7 | **0.75** | **0.47** | **0.58** | **0.73** | **0.3** | **0.43** |

**Table 1.** Macro and micro-level evaluation results across both models and all seven prompts on HPO-GS

| GPT version | Prompt | Precision | Recall | F1 | Precision | Recall | F1 |
|---|---|---|---|---|---|---|---|
| | | **Macro-level** | | | **Micro-level** | | |
| 3.5 | 1 | 0.51 | 0.12 | 0.19 | 0.5 | 0.11 | 0.18 |
| | 2 | **0.68** | 0.05 | 0.09 | **0.68** | 0.05 | 0.09 |
| | 3 | 0.27 | 0.29 | 0.28 | 0.26 | 0.25 | 0.25 |
| | 4 | 0.26 | 0.33 | 0.29 | 0.22 | 0.29 | 0.25 |
| | 5 | 0.31 | 0.2 | 0.24 | 0.3 | 0.17 | 0.22 |
| | 6 | 0.31 | 0.2 | 0.24 | 0.3 | 0.17 | 0.22 |
|   | 7 | 0.56 | **0.56** | **0.56** | 0.54 | **0.49** | **0.51** |
| 4 | 1 | 0.46 | 0.44 | 0.45 | 0.45 | 0.39 | 0.42 |
|   | 2 | 0.44 | 0.44 | 0.44 | 0.43 | 0.38 | 0.4 |
|   | 3 | 0.47 | 0.43 | 0.45 | 0.47 | 0.37 | 0.41 |
|   | 4 | 0.43 | 0.53 | 0.47 | 0.43 | 0.46 | 0.44 |
|   | 5 | 0.44 | 0.27 | 0.33 | 0.43 | 0.24 | 0.31 |
|   | 6 | 0.44 | 0.27 | 0.33 | 0.43 | 0.24 | 0.31 |
|   | 7 | **0.78** | **0.73** | **0.75** | **0.77** | **0.64** | **0.7** |

**Table 2.** Macro and micro-level evaluation results across both models and all seven prompts on BIOC-GS

Tables 1 and 2 list the experimental results achieved by both models across all seven prompts on HPO-GS and BIOC-GS respectively, while Table 3 lists, as reference point, the results of the state of the art methods for phenotype concept recognition. Below we discuss the main findings emerging from these results:

- *The few-shot learning strategy achieves results comparable or better than the state of the art*. Phenotype concept recognition is known to be a difficult task - as showcased by the F1 scores listed in Table 3, which are roughly 0.2 lower that other domain-specific concept recognition tasks, such as gene or drug names. The GPT models perform significantly lower that the state of the art in most cases, with the micro-level evaluation F1 scores being half the values of tools such as PhenoTagger or the Monarch Annotator (note that the latter does not rely on a BERT-based architecture, or in general on a LLM-based on neural network-based architecture). The few-shot learning strategy, however, showcases the power of generative models. While on HPO-GS the results are comparable to the state of the art, on BIOC-GS gpt-4.0 surpasses the best in class with a significant margin – almost 0.1 (0.7 F1 on micro-level evaluation compared to 0.61 F1 for PhenoTagger). Although the setup employed for this strategy would not serve phenotype concept recognition in general, it would support manual annotation in a clearly defined domain - e.g., cardiovascular diseases. Additional experiments are described in the following section.
- *Both models have a consistent behaviour across prompts*. Prompts 1 and 2 - defining the task as an extraction of phenotypes and clinical abnormalities, followed by an alignment to HPO IDs - achieve the best precision, with the increased focus of Prompt 2 leading to better results when using gpt-3.5 (although this change had no effect on gpt-4.0). Similarly, Prompt 7 (few-shot learning) achieved the best recall - which was expected since the examples included all concepts present in the gold standard.
- *Macro and micro-level evaluation results show significant discrepancies*. While some differences were expected, the micro-level experimental results were surprisingly lower than the macro-level results on HPO-GS. This could be attributed to the variability of the lexical representations of the concepts in text - e.g., *Brachydactyly C* vs *Brachydactyly, type C.* This outcome does not hold on BIOC-GS, which seems to be more uniform.

|   | **HPO-GS** | | | **BioC-GS** | | |
|---|---|---|---|---|---|---|
| Tool | **Precision** | **Recall** | **F1** | **Precision** | **Recall** | **F1** |
| PhenoTagger [25] | 0.77 | **0.52** | **0.62** | 0.74 | 0.52 | **0.61** |

| | | | | | |
|---|---|---|---|---|---|
| ClinPheno [26] | 0.73 | 0.30 | 0.43 | 0.47 | **0.57** | 0.52 |
| Doc2HPO [27] | 0.81 | 0.41 | 0.55 | **0.84** | 0.29 | 0.43 |
| Monarch Annotator [9] | **0.82** | 0.50 | **0.62** | 0.47 | 0.46 | 0.46 |
| NCBO Annotator [28] | 0.68 | 0.48 | 0.57 | 0.78 | 0.41 | 0.54 |

**Table 3.** Micro-level evaluation results of the state of the art methods for phenotype concept recognition

**Hallucinations**

Hallucinations represent the generation of inaccurate, nonsensical, or text irrelevant to the given context. Our experiments defined standard, community-accepted phenotype concept recognition tasks and the evaluation targeted HPO IDs extracted by the models. Hence, in terms of hallucinations, the expectation was to find non-existing HPO IDs in the output produced by the models. Table 3 lists an overview of the number of concepts (identified by HPO IDs) extracted in our experiments, in addition to the number of hallucinations. It can be observed that the latter has insignificant levels, and as such, hallucinations do not pose challenges for this task. Some examples of hallucinations include: HP:0020115, HP:0025111, HP:0023656, HP:0031966, HP:0020019, HP:0040060. A second observation can be made with respect to Prompts 3 and 4 (instructing the model to perform the task by its name): these prompts are very prolific on HPO-GS (7780 HPO IDs found, and 6698, respectively), which leads to an increased recall and a lower precision.

| Model | Prompt | Total found | Unique | Hall's | Hall's (%) | Total found | Unique | Hall's | Hall's (%) |
|---|---|---|---|---|---|---|---|---|---|
| | | | | | | | | | |
| **BASE** | | **2,773** | **497** | | | 783 | 358 | | |
| | | | | | | | | | |
| 3.5 | 1 | 978 | 408 | 0 | 0 | 167 | 126 | 0 | 0 |
| | 2 | 460 | 237 | 0 | 0 | 53 | 45 | 0 | 0 |
| | 3 | 7780 | 1546 | 11 | 1 | 755 | 432 | 0 | 0 |
| | 4 | 6698 | 1980 | 14 | 1 | 996 | 484 | 0 | 0 |
| | 5 | 1095 | 841 | 3 | 0 | 351 | 242 | 0 | 0 |
| | 6 | 771 | 491 | 4 | 1 | 442 | 316 | 0 | 0 |
| | 7 | 2551 | 681 | 2 | 0 | 709 | 319 | 1 | 0 |
| 4 | 1 | 1617 | 634 | 1 | 0 | 666 | 390 | 0 | 0 |
| | 2 | 1605 | 633 | 2 | 0 | 699 | 397 | 0 | 0 |
| | 3 | 1534 | 728 | 2 | 0 | 625 | 366 | 1 | 0 |
| | 4 | 2003 | 855 | 7 | 1 | 839 | 479 | 3 | 1 |
| | 5 | 1408 | 636 | 4 | 1 | 1287 | 648 | 3 | 0 |

|   | 6 | 977 | 483 | 4 | 1 | 432 | 288 | 2 | 1 |
|   | 7 | 3469 | 938 | 0 | 0 | 676 | 313 | 1 | 0 |

Table 4. Overview of number of HPO IDs found in all experiments and associated hallucinations

**Few-shot learning with different sets of concepts**

The results achieved by Prompt 7 and discussed above relied on the same set of concepts as those present in the gold standard. To test the impact of this set on the results (which in a standard setting would be expected, since the entire ontology would be considered), we performed three additional experiments using gpt-4.0 and Prompt 7. Firstly, we used the concepts covered by HPO-GS to do few-shot learning for BIOC-GS. The results were significantly lower, the model achieving macro-level precision, recall and F1 of 0.25, 0.23, 0.24 respectively and micro-level metrics of 0.23, 0.2, 0.21.

Secondly, we used the top-level profile of the two corpora (depicted in Fig. 3; i.e., the majority of the concepts describing musculo-skeletal abnormalities) to generate a random set of ~1,000 concepts. This resulted in a set comprising 1,165 HPO concepts (~43KB in size with labels and ~7% of the entire ontology) and the following overlaps with the two gold standard corpora: (i) 160 concepts overlap with BIOC-GS – i.e., 45% of BIOC-GS and 14% of the learning set; (ii) 138 concepts overlap with HPO-GS – i.e., 30% of HPO-GS and 12% of the learning set. We re-ran Prompt 7 on gpt-4.0 on both corpora and the results – as shown in Table 5 – are encouraging.

|         | **Macro-level** | | | **Micro-level** | | |
|---------|-----------|--------|------|-----------|--------|------|
|         | Precision | Recall | F1   | Precision | Recall | F1   |
| HPO-GS  | 0.63      | 0.29   | 0.4  | 0.6       | 0.19   | 0.29 |
| BIOC-GS | 0.55      | 0.42   | 0.48 | 0.5       | 0.37   | 0.43 |

Table 5. Experimental results on using a random set of concepts for few-shot learning

These results support our assumption that gpt-4.0 would be useful for annotation purposes in a domain-specific setting, without the need to use the entire set of concepts describing the domain to perform few-shot learning.

**Concordance across prompts**

A complete overview of the pairwise concordance of the outcomes across both models and all prompts is provided in Appendix 2 and 3 in the Supplementary material. More concretely, we recorded the percentage of common correct and incorrect HPO IDs when considering one model output as base reference. For example, on HPO-GS 51% of the correct HPO IDs extracted by gpt-3.5 Prompt 1 are in common with the correct HPO IDs extracted by gpt-3.5 Prompt 2, with this common set representing 93% of the total correct HPO IDs extracted by the latter.

Overall, the results vary significantly and there is no combination of model - prompt that achieved a high level of agreement on both correctly and incorrectly extracted HPO IDs. A stand-out is perhaps gpt-3.5 Prompt 2 that achieves a rather consistent level of agreement with most of the other prompts on both corpora: (i) on HPO-GS - 93% correct in common with Prompt 1 - which is expected because Prompt 2 targets conceptually a subset of Prompt 1, 87% with Prompt 7, 84% with gpt-4.0 Prompt 1; (ii) on BIOC-GS – 97% correct in common with Prompt 1, 81% with Prompt 7 and over 80% with all gpt-4.0 experiments except for Prompts 5 and 6.

Appendix 4 in the Supplementary material lists the top 5 incorrectly extracted HPO IDs across all experiments. These HPO IDs are fairly consistent within the context of a model and completely divergent across models. For example, the most common errors of gpt-3.5 are: *Decreased body*

*weight* (HP:0004325), *Intellectual disability, profound* (HP:0002187), *Joint hypermobility* (HP:0001382), *Abnormality of the nervous system* (HP:0000707), while those of gpt-4.0 are: *Poor wound healing* (HP:0001058), *Cerebral hamartoma* (HP:0009731).

It is interesting to note the nature of failures in concept mapping. For example, gpt-3.5 tags the text 'Angelman's syndrome' (a disease not present in HPO) with *Decreased body weight* (HP:0004325 – shown in Fig. 1), and 'Prader-Willi syndrome' (another disease not present in HPO), 'bilateral acoustic neuromas' or 'Neurofibromatosis type 2' with *Intellectual disability, profound* (HP:0002187), while gpt-4.0 tags, consistently, the same text spans, e.g., 'Angelman's syndrome', with *Poor wound healing* (HP:0001058) or 'Neurofibromatosis type 2' with *Cerebral hamartoma* (HP:0009731).

**Same model and prompt concordance**

A final experiment was performed to understand the concordance across different runs of the same model and prompt. We ran five times the annotation experiment using gpt-4.0 Prompt 1. Overall, all runs achieved the same precision and recall, with very minor differences (+/- 0.01). The concordance in the results produced by the runs was, however, surprisingly low. Across all runs, we found the common set of: 75.82% of all correctly identified HPO IDs; 28.09% of all incorrectly identified HPO IDs, and 86.6% of all concepts not found by the models. This shows a high level of divergence in concept mapping errors produced by the individual runs.

**LIMITATIONS**

A summary of the limitations derived from the experiments discussed above is listed below:
- The few-shot learning strategy adopted in our experiments - while surpassing the state of the art in some cases - defeats the general purpose of open phenotype concept recognition. Due to limitations in the size of the input data, we restricted the examples to only the concepts present in the gold standard. In a real-world scenario, this set of concepts is unknown - and hence this strategy would fail. Our experiments did, however, show that ontology stratification strategies could be employed as an alternative to using the entire ontology – e.g., domain-specific selection. Cost is still a prohibitive factor for this approach. For example, the few-shot learning experiments on BIOC-GS costed USD $50, which used only 454 entries with an average length of 56 characters + the learning component of 358 unique ontology concepts (~12KB)
- The performance of the model is non-deterministic. Executing the same prompt over the same input leads to slightly different results. This is particularly challenging as it hinders the establishment of an accurate ground truth and leaves a degree of uncertainty in completeness always associated with the outcomes.
- The choice of wording in the prompt influences the results. While this is expected (hence the need for iterative prompt engineering), it is also remarkably challenging when considering the lack of concordance between the outcomes - as shown in Appendix 2 and 3 in the Supplementary material (e.g., prompts that have been iterated on produce HPO IDs that are not found by subsequent prompts)

**CONCLUSION**

This paper presents a study that assesses the capabilities of the GPT models underpinning ChatGPT to perform phenotype concept recognition, using concepts grounded in the Human Phenotype Ontology, assuming a need for manual curation / annotation of publications or clinical records. The experimental setup covered both gpt-3.5 and gpt-4.0 and a series of seven prompts ranging from direct instructions to perform the task by name to chains of named entity recognition followed by concept mapping and to few-shot learning. The results show that with an appropriate set-up – in this case few-shot learning – these models can surpass the best-in-class tools, which are either using BERT-based architectures or more classical natural language processing pipelines.

The main challenges to a direct adoption of these models, document by our error analysis include the non-deterministic outputs of the models, the lack of concordance between different prompt outputs, as well as between different runs with the same prompt. Unlike other use cases, hallucinations do not affect the task we have focused on.

**COMPETING INTERESTS**

None.

**FUNDING**

GB is supported by the Angela Wright Bennett Foundation, The Stan Perron Charitable Foundation, the McCusker Charitable Foundation via Channel 7 Telethon Trust and Mineral Resources via the Perth Children's Hospital Foundation. CJM, JHC, and JR are funded by NIH HG010860, NIH OD011883 and the Director, Office of Science, Office of Basic Energy Sciences, of the US Department of Energy DE-AC0205CH1123. MAH and PNR are funded by NIH OD011883, NHGRI RM1HG010860, and NHGRI U24HG011449

# APPENDIX 1: Prompt 7 example (PMID: 292745)

Examples: The Human Phenotype Ontology defines phenotype concepts using the following label – HPO ID associations:
Hypospadias // HP:0000047
Ureteral duplication // HP:0000073
Abnormality of the kidney // HP:0000077
Abnormality of the genital system // HP:0000078
Abnormality of the urinary system // HP:0000079
Duplicated collecting system // HP:0000081
Ectopic kidney // HP:0000086
Renal hypoplasia // HP:0000089
Renal agenesis // HP:0000104
Renal dysplasia // HP:0000110
Polycystic kidney dysplasia // HP:0000113
Unilateral renal agenesis // HP:0000122
Abnormality of the ovary // HP:0000137
Rectovaginal fistula // HP:0000143
Abnormality of the mouth // HP:0000153
Wide mouth // HP:0000154
Abnormal lip morphology // HP:0000159
Abnormality of the dentition // HP:0000164
Abnormality of the gingiva // HP:0000168
Microglossia // HP:0000171
Abnormal palate morphology // HP:0000174
Cleft palate // HP:0000175
Abnormal upper lip morphology // HP:0000177
Movement abnormality of the tongue // HP:0000182
Lower lip pit // HP:0000196
Orofacial cleft // HP:0000202
Trismus // HP:0000211
Hydrocephalus // HP:0000238
Parietal bossing // HP:0000242
Brachycephaly // HP:0000248
Microcephaly // HP:0000252
Macrocephaly // HP:0000256
Abnormality of the face // HP:0000271
Malar flattening // HP:0000272
Abnormal mandible morphology // HP:0000277
Coarse facial features // HP:0000280
Epicanthus // HP:0000286
Abnormality of the philtrum // HP:0000288
Mandibular prognathia // HP:0000303
Pointed chin // HP:0000307
Microretrognathia // HP:0000308
Hypertelorism // HP:0000316
Short philtrum // HP:0000322
Micrognathia // HP:0000347
Strabismus // HP:0000486
Deeply set eye // HP:0000490
Downslanted palpebral fissures // HP:0000494
Abnormality of eye movement // HP:0000496
Ptosis // HP:0000508
Cataract // HP:0000518
Subcapsular cataract // HP:0000523

Abnormality iris morphology // HP:0000525
Myopia // HP:0000545
Keratoconus // HP:0000563
Microphthalmia // HP:0000568
Exotropia // HP:0000577
Nasolacrimal duct obstruction // HP:0000579
Coloboma // HP:0000589
Blindness // HP:0000618
Nystagmus // HP:0000639
Widely spaced teeth // HP:0000687
Atypical behavior // HP:0000708
Autism // HP:0000717
Aggressive behavior // HP:0000718
Autistic behavior // HP:0000729
Inappropriate laughter // HP:0000748
Paroxysmal bursts of laughter // HP:0000749
Hyperactivity // HP:0000752
Abnormal peripheral nervous system morphology // HP:0000759
Peripheral axonal degeneration // HP:0000764
Abnormal sternum morphology // HP:0000766
Pectus excavatum // HP:0000767
Abnormal rib morphology // HP:0000772
Hyperparathyroidism // HP:0000843
Abnormality of the outer ear // HP:0000356
Abnormality of the inner ear // HP:0000359
Tinnitus // HP:0000360
Hearing abnormality // HP:0000364
Hearing impairment // HP:0000365
Abnormality of the nose // HP:0000366
Low-set ears // HP:0000369
Abnormality of the middle ear // HP:0000370
Abnormal cochlea morphology // HP:0000375
Incomplete partition of the cochlea type II // HP:0000376
Abnormal pinna morphology // HP:0000377
Cupped ear // HP:0000378
Stapes ankylosis // HP:0000381
Abnormal periauricular region morphology // HP:0000383
Preauricular skin tag // HP:0000384
Lop ear // HP:0000394
Recurrent otitis media // HP:0000403
Conductive hearing impairment // HP:0000405
Sensorineural hearing impairment // HP:0000407
Mixed hearing impairment // HP:0000410
Atresia of the external auditory canal // HP:0000413
Wide nasal bridge // HP:0000431
Depressed nasal tip // HP:0000437
Abnormality of the neck // HP:0000464
Abnormality of the skin // HP:0000951
Hyperpigmentation of the skin // HP:0000953
Cafe-au-lait spot // HP:0000957
Edema // HP:0000969
Hypopigmentation of the skin // HP:0001010
Multiple lipomas // HP:0001012
Albinism // HP:0001022
Numerous nevi // HP:0001054

Milia // HP:0001056
Pterygium // HP:0001059
Neurofibroma // HP:0001067
Ocular albinism // HP:0001107
Limbal dermoid // HP:0001140
Orbital cyst // HP:0001144
Abnormality of the hand // HP:0001155
Brachydactyly // HP:0001156
Syndactyly // HP:0001159
Arachnodactyly // HP:0001166
Abnormal thumb morphology // HP:0001172
Preaxial hand polydactyly // HP:0001177
Hand clenching // HP:0001188
Ulnar deviation of the hand or of fingers of the hand // HP:0001193
Triphalangeal thumb // HP:0001199
Distal symphalangism of hands // HP:0001204
Failure to thrive // HP:0001508
Growth delay // HP:0001510
Intrauterine growth retardation // HP:0001511
Obesity // HP:0001513
Anteriorly placed anus // HP:0001545
Overgrowth // HP:0001548
Oligohydramnios // HP:0001562
Abnormality of the nail // HP:0001597
Hypernasal speech // HP:0001611
Premature birth // HP:0001622
Abnormal heart morphology // HP:0001627
Truncus arteriosus // HP:0001660
Bradycardia // HP:0001662
Abnormal foot morphology // HP:0001760
Talipes equinovarus // HP:0001762
Toe syndactyly // HP:0001770
Bilateral talipes equinovarus // HP:0001776
Small nail // HP:0001792
Anonychia // HP:0001798
Rocker bottom foot // HP:0001838
Talipes // HP:0001883
Abnormal clavicle morphology // HP:0000889
Cervical ribs // HP:0000891
Bifid ribs // HP:0000892
Sprengel anomaly // HP:0000912
Abnormality of the skeletal system // HP:0000924
Abnormality of the vertebral column // HP:0000925
Abnormal facial shape // HP:0001999
Frontal bossing // HP:0002007
Potter facies // HP:0002009
Abnormality of the abdominal organs // HP:0002012
Vomiting // HP:0002013
Dysphagia // HP:0002015
Anal atresia // HP:0002023
Abdominal pain // HP:0002027
Abnormal rectum morphology // HP:0002034
Cerebral atrophy // HP:0002059
Abnormal cerebral morphology // HP:0002060
Gait ataxia // HP:0002066

Bilateral tonic-clonic seizure // HP:0002069
Limb ataxia // HP:0002070
Abnormal lung morphology // HP:0002088
Pulmonary hypoplasia // HP:0002089
Pneumothorax // HP:0002107
Abnormal cerebral ventricle morphology // HP:0002118
Generalized myoclonic seizure // HP:0002123
Status epilepticus // HP:0002133
Broad-based gait // HP:0002136
Abnormality of speech or vocalization // HP:0002167
Intellectual disability, profound // HP:0002187
Dysgenesis of the cerebellar vermis // HP:0002195
White forelock // HP:0002211
Premature graying of hair // HP:0002216
Poor motor coordination // HP:0002275
Global brain atrophy // HP:0002283
Incoordination // HP:0002311
Vertigo // HP:0002321
Drowsiness // HP:0002329
Delayed ossification of carpal bones // HP:0001216
Intellectual disability // HP:0001249
Seizure // HP:0001250
Ataxia // HP:0001251
Hypotonia // HP:0001252
Intellectual disability, mild // HP:0001256
Spasticity // HP:0001257
Global developmental delay // HP:0001263
Hemiparesis // HP:0001269
Motor delay // HP:0001270
Cerebellar atrophy // HP:0001272
Gait disturbance // HP:0001288
Abnormal cranial nerve morphology // HP:0001291
Encephalopathy // HP:0001298
Abnormal cerebellum morphology // HP:0001317
Neonatal hypotonia // HP:0001319
Muscle weakness // HP:0001324
Specific learning disability // HP:0001328
Myoclonus // HP:0001336
Absent speech // HP:0001344
Hyperreflexia // HP:0001347
Craniosynostosis // HP:0001363
Abnormal joint morphology // HP:0001367
Flexion contracture // HP:0001371
Hip dysplasia // HP:0001385
Cerebral calcification // HP:0002514
Increased intracranial pressure // HP:0002516
Inability to walk // HP:0002540
Episodic abdominal pain // HP:0002574
Colitis // HP:0002583
Polyphagia // HP:0002591
Hypotension // HP:0002615
Madelung deformity // HP:0003067
Defective DNA repair after ultraviolet radiation damage // HP:0003079
Limb joint contracture // HP:0003121
Skeletal muscle atrophy // HP:0003202

Prominent scrotal raphe // HP:0003246
Spina bifida occulta // HP:0003298
Spondylolisthesis // HP:0003302
Spondylolysis // HP:0003304
Spinal canal stenosis // HP:0003416
Abnormal 2nd finger morphology // HP:0004100
Short middle phalanx of the 5th finger // HP:0004220
Abnormal gastric mucosa morphology // HP:0004295
Abnormal renal collecting system morphology // HP:0004742
Rectoperineal fistula // HP:0004792
Absence of the pulmonary valve // HP:0005134
Internal carotid artery hypoplasia // HP:0005290
Bridged sella turcica // HP:0005449
EEG abnormality // HP:0002353
Sleep disturbance // HP:0002360
Abnormal brainstem morphology // HP:0002363
Spina bifida // HP:0002414
Language impairment // HP:0002463
Hyperkinetic movements // HP:0002487
Absent trapezoid bone // HP:0006106
Abnormal finger flexion crease // HP:0006143
Proximal symphalangism of hands // HP:0006152
Calcification of falx cerebri // HP:0005462
Fusion of middle ear ossicles // HP:0005473
Secondary microcephaly // HP:0005484
Duplication of renal pelvis // HP:0005580
Hypopigmentation of hair // HP:0005599
Positional foot deformity // HP:0005656
Distal arthrogryposis // HP:0005684
Short middle phalanx of finger // HP:0005819
Biliary atresia // HP:0005912
Multiple pulmonary cysts // HP:0005948
Paraspinal neurofibroma // HP:0006751
Dysplastic patella // HP:0006446
Proximal femoral epiphysiolysis // HP:0006461
Abnormality of the alveolar ridges // HP:0006477
Cerebellar calcifications // HP:0007352
Focal-onset seizure // HP:0007359
Confetti-like hypopigmented macules // HP:0007449
Generalized hypopigmentation // HP:0007513
Bilateral microphthalmos // HP:0007633
Lacrimal duct stenosis // HP:0007678
Vitreoretinopathy // HP:0007773
Posterior subcapsular cataract // HP:0007787
Scoliosis // HP:0002650
Skeletal dysplasia // HP:0002652
Basal cell carcinoma // HP:0002671
Large foramen magnum // HP:0002700
Bilateral cleft lip and palate // HP:0002744
Congenital contracture // HP:0002803
Kyphosis // HP:0002808
Abnormality of limb bone morphology // HP:0002813
Abnormality of the lower limb // HP:0002814
Abnormality of the upper limb // HP:0002817
Abnormal morphology of the radius // HP:0002818

Multiple joint contractures // HP:0002828
Meningioma // HP:0002858
Medulloblastoma // HP:0002885
Ependymoma // HP:0002888
Hemivertebrae // HP:0002937
Vertebral fusion // HP:0002948
Lacrimal duct aplasia // HP:0007925
Malformed lacrimal duct // HP:0007993
Peripheral opacification of the cornea // HP:0008011
Neoplasm of the skin // HP:0008069
Ankylosis of feet small joints // HP:0008090
Intervertebral disc degeneration // HP:0008419
Cochlear malformation // HP:0008554
Hypoplasia of the cochlea // HP:0008586
Supraauricular pit // HP:0008606
Bilateral sensorineural hearing impairment // HP:0008619
Nonprogressive encephalopathy // HP:0007030
Atypical absence seizure // HP:0007270
Unilateral vestibular schwannoma // HP:0009590
Abnormality of the vestibulocochlear nerve // HP:0009591
Astrocytoma // HP:0009592
Peripheral schwannoma // HP:0009593
Retinal hamartoma // HP:0009594
Carpal synostosis // HP:0009702
Adenoma sebaceum // HP:0009720
Rhabdomyoma // HP:0009730
Glioma // HP:0009733
Spinal neurofibroma // HP:0009735
Lisch nodules // HP:0009737
Ankyloblepharon // HP:0009755
Popliteal pterygium // HP:0009756
Short thumb // HP:0009778
Branchial anomaly // HP:0009794
Branchial fistula // HP:0009795
Branchial cyst // HP:0009796
Cholesteatoma // HP:0009797
Peripheral neuropathy // HP:0009830
Mononeuropathy // HP:0009831
Aplasia/Hypoplasia of the middle phalanges of the hand // HP:0009843
Aplasia of the distal phalanges of the hand // HP:0009881
Abnormal temporal bone morphology // HP:0009911
Duane anomaly // HP:0009921
Partial duplication of thumb phalanx // HP:0009944
Type A brachydactyly // HP:0009370
Type A1 brachydactyly // HP:0009371
Type A2 brachydactyly // HP:0009372
Type C brachydactyly // HP:0009373
Fever // HP:0001945
Abnormal scalp morphology // HP:0001965
Short 4th metacarpal // HP:0010044
Short 5th metacarpal // HP:0010047
Short hallux // HP:0010109
Pseudoepiphyses of the phalanges of the hand // HP:0010235
Ulnar deviation of the 2nd finger // HP:0009464
Ulnar deviation of finger // HP:0009465

Pseudoepiphysis of the 2nd finger // HP:0009495
Short 2nd finger // HP:0009536
Vestibular schwannoma // HP:0009588
Bilateral vestibular schwannoma // HP:0009589
Polydactyly // HP:0010442
Long toe // HP:0010511
Coronal hypospadias // HP:0008743
Feeding difficulties in infancy // HP:0008872
Postnatal growth retardation // HP:0008897
Generalized neonatal hypotonia // HP:0008935
Abnormal axial skeleton morphology // HP:0009121
Protruding tongue // HP:0010808
Atonic seizure // HP:0010819
Mild global developmental delay // HP:0011342
Moderate global developmental delay // HP:0011343
Localized skin lesion // HP:0011355
Morphological abnormality of the semicircular canal // HP:0011380
Abnormality of the incus // HP:0011453
Feeding difficulties // HP:0011968
Camptodactyly // HP:0012385
Pain // HP:0012531
Bilateral renal dysplasia // HP:0012582
Involuntary movements // HP:0004305
Ventricular arrhythmia // HP:0004308
Short stature // HP:0004322
Abnormality of bone mineral density // HP:0004348
Abnormal circulating calcium concentration // HP:0004363
Neoplasm of the nervous system // HP:0004375
Abnormality of the anus // HP:0004378
Abnormality of the middle ear ossicles // HP:0004452
Dilatated internal auditory canal // HP:0004458
Postauricular pit // HP:0004464
Preauricular pit // HP:0004467
Relative macrocephaly // HP:0004482
Clinodactyly // HP:0030084
Spinal cord tumor // HP:0010302
Profound global developmental delay // HP:0012736
Papilloma // HP:0012740
Neurodevelopmental abnormality // HP:0012759
Overlapping fingers // HP:0010557
Odontogenic keratocysts of the jaw // HP:0010603
Chalazion // HP:0010605
Skin tags // HP:0010609
Palmar pits // HP:0010610
Plantar pits // HP:0010612
Angiofibromas // HP:0010615
Cardiac fibroma // HP:0010617
Ovarian fibroma // HP:0010618
Ectopic calcification // HP:0010766
Erythema // HP:0010783
Happy demeanor // HP:0040082
Premature skin wrinkling // HP:0100678
Abnormality of the seventh cranial nerve // HP:0010827
EEG with persistent abnormal rhythmic activity // HP:0010846
EEG with spike-wave complexes (2.5-3.5 Hz) // HP:0010848

Intellectual disability, severe // HP:0010864
Bilateral renal agenesis // HP:0010958
Epileptic spasm // HP:0011097
Papule // HP:0200034
Epidermoid cyst // HP:0200040
Malignant mesothelioma // HP:0100001
Neoplasm of the central nervous system // HP:0100006
Neoplasm of the peripheral nervous system // HP:0100007
Schwannoma // HP:0100008
Intracranial meningioma // HP:0100009
Spinal meningioma // HP:0100010
Scleral schwannoma // HP:0100011
Epiretinal membrane // HP:0100014
Capsular cataract // HP:0100017
Cortical cataract // HP:0100019
Posterior capsular cataract // HP:0100020
Cerebral palsy // HP:0100021
Abnormality of movement // HP:0100022
Recurrent hand flapping // HP:0100023
Conspicuously happy disposition // HP:0100024
Sarcoma // HP:0100242
Fibrosarcoma // HP:0100244
Ectrodactyly // HP:0100257
Preaxial polydactyly // HP:0100258
Proximal symphalangism // HP:0100264
Paramedian lip pit // HP:0100269
Branchial sinus // HP:0100272
Gustatory lacrimation // HP:0100274
Ulcerative colitis // HP:0100279
Unilateral cleft lip // HP:0100333
Unilateral cleft palate // HP:0100334
Bilateral cleft lip // HP:0100336
Neoplasm of the endocrine system // HP:0100568
Short ear // HP:0400005

Task: Using the list above, find Human Phenotype Ontology concepts in the following text and return their associated IDs for every appearance in the text: This paper is based on our experience with the Gorlin-Goltz syndrome and on data from 14 patients of the Nordwestdeutsche Kieferklinik in whom this disorder was detected, treated and followed up. A clinical concept has been produced, with a diagnostic check list including a genetic and a dermatological routine work up as well as a radiological survey of the jaws and skeleton. Whenever multiple basal cell carcinomas plus the typical jaw lesions are found in a patient, the diagnosis is easy. A minimum diagnostic criterion is the combination of either the skin tumours or multiple odontogenic keratocysts plus a positive family history for this disorder, bifid ribs, lamellar calcification of the falx cerebri or any one of the skeletal abnormalities typical of this syndrome. All those in whom this disorder is diagnosed or suspected should be followed up for the rest of their lives. The family should be examined and genetic counselling should be offered.

# APPENDIX 2: Pairwise concordance across all models and prompts using the HPO-GS corpus

**Base:** gpt-3.5, Prompt 1

| Model | Prompt | Common correct in base prompt | Common correct in comparing prompt | Common incorrect in base prompt | Common incorrect in comparing prompt |
|---|---|---|---|---|---|
| 3.5 | 2 | 0.51 | 0.93 | 0.27 | 0.59 |
| | 3 | 0.57 | 0.51 | 0.21 | 0.03 |
| | 5 | 0.25 | 0.59 | 0.1 | 0.05 |
| | 6 | 0.37 | 0.63 | 0.12 | 0.1 |
| | 4 | 0.62 | 0.5 | 0.29 | 0.04 |
| | 7 | 0.79 | 0.43 | 0.14 | 0.06 |
| 4.0 | 1 | 0.77 | 0.5 | 0.09 | 0.04 |
| | 2 | 0.77 | 0.52 | 0.09 | 0.05 |
| | 3 | 0.72 | 0.53 | 0.08 | 0.04 |
| | 5 | 0.5 | 0.51 | 0.06 | 0.03 |
| | 6 | 0.38 | 0.52 | 0.04 | 0.03 |
| | 4 | 0.77 | 0.46 | 0.1 | 0.03 |
| | 7 | 0.8 | 0.4 | 0.09 | 0.02 |

**Base:** gpt-3.5, Prompt 2

| Model | Prompt | Common correct in base prompt | Common correct in comparing prompt | Common incorrect in base prompt | Common incorrect in comparing prompt |
|---|---|---|---|---|---|
| 3.5 | 1 | 0.93 | 0.51 | 0.59 | 0.27 |
| | 3 | 0.64 | 0.31 | 0.18 | 0.01 |
| | 5 | 0.25 | 0.33 | 0.08 | 0.02 |
| | 6 | 0.44 | 0.42 | 0.12 | 0.05 |
| | 4 | 0.65 | 0.29 | 0.34 | 0.02 |
| | 7 | 0.87 | 0.26 | 0.2 | 0.04 |
| 4.0 | 1 | 0.84 | 0.3 | 0.13 | 0.03 |
| | 2 | 0.85 | 0.31 | 0.13 | 0.03 |
| | 3 | 0.82 | 0.33 | 0.13 | 0.03 |
| | 5 | 0.59 | 0.33 | 0.1 | 0.02 |
| | 6 | 0.49 | 0.36 | 0.05 | 0.02 |
| | 4 | 0.86 | 0.28 | 0.14 | 0.02 |
| | 7 | 0.85 | 0.24 | 0.13 | 0.01 |

**Base:** gpt-3.5, Prompt 3

| Model | Prompt | Common correct in base prompt | Common correct in comparing prompt | Common incorrect in base prompt | Common incorrect in comparing prompt |
|---|---|---|---|---|---|
| 3.5 | 1 | 0.51 | 0.57 | 0.03 | 0.21 |
| | 2 | 0.31 | 0.64 | 0.01 | 0.18 |
| | 5 | 0.22 | 0.57 | 0.02 | 0.06 |
| | 6 | 0.32 | 0.61 | 0.03 | 0.16 |
| | 4 | 0.63 | 0.57 | 0.13 | 0.11 |
| | 7 | 0.69 | 0.42 | 0.04 | 0.1 |
| 4.0 | 1 | 0.65 | 0.48 | 0.01 | 0.04 |
| | 2 | 0.65 | 0.48 | 0.01 | 0.05 |
| | 3 | 0.59 | 0.48 | 0.01 | 0.05 |
| | 5 | 0.39 | 0.45 | 0.01 | 0.03 |
| | 6 | 0.32 | 0.48 | 0.01 | 0.03 |
| | 4 | 0.65 | 0.43 | 0.02 | 0.04 |
| | 7 | 0.65 | 0.36 | 0.02 | 0.03 |

**Base:** gpt-3.5, Prompt 4

| Model | Prompt | Common correct in base prompt | Common correct in comparing prompt | Common incorrect in base prompt | Common incorrect in comparing prompt |
|---|---|---|---|---|---|
| 3.5 | 1 | 0.5 | 0.62 | 0.04 | 0.29 |
| | 2 | 0.29 | 0.65 | 0.02 | 0.34 |
| | 3 | 0.57 | 0.63 | 0.11 | 0.13 |
| | 5 | 0.21 | 0.61 | 0.02 | 0.07 |
| | 6 | 0.29 | 0.62 | 0.02 | 0.16 |
| | 7 | 0.68 | 0.46 | 0.04 | 0.12 |
| 4.0 | 1 | 0.63 | 0.51 | 0.01 | 0.05 |
| | 2 | 0.62 | 0.51 | 0.01 | 0.05 |
| | 3 | 0.56 | 0.51 | 0.02 | 0.06 |
| | 5 | 0.41 | 0.52 | 0.01 | 0.02 |
| | 6 | 0.3 | 0.5 | 0.01 | 0.03 |
| | 4 | 0.63 | 0.47 | 0.01 | 0.03 |
| | 7 | 0.65 | 0.41 | 0.03 | 0.04 |

**Base:** gpt-3.5, Prompt 5

| Model | Prompt | Common correct in base prompt | Common correct in comparing prompt | Common incorrect in base prompt | Common incorrect in comparing prompt |
|---|---|---|---|---|---|
| 3.5 | 1 | 0.59 | 0.25 | 0.05 | 0.1 |
| | 2 | 0.33 | 0.25 | 0.02 | 0.08 |
| | 3 | 0.57 | 0.22 | 0.06 | 0.02 |
| | 6 | 0.54 | 0.4 | 0.08 | 0.14 |
| | 4 | 0.61 | 0.21 | 0.07 | 0.02 |
| | 7 | 0.71 | 0.17 | 0.03 | 0.03 |
| 4.0 | 1 | 0.64 | 0.18 | 0.02 | 0.02 |
| | 2 | 0.63 | 0.18 | 0.02 | 0.02 |
| | 3 | 0.58 | 0.18 | 0.02 | 0.02 |
| | 5 | 0.42 | 0.18 | 0.01 | 0.01 |
| | 6 | 0.34 | 0.19 | 0 | 0.01 |
| | 4 | 0.67 | 0.17 | 0.02 | 0.01 |
| | 7 | 0.68 | 0.15 | 0.01 | 0 |

**Base:** gpt-3.5, Prompt 6

| Model | Prompt | Common correct in base prompt | Common correct in comparing prompt | Common incorrect in base prompt | Common incorrect in comparing prompt |
|---|---|---|---|---|---|
| 3.5 | 1 | 0.63 | 0.37 | 0.1 | 0.12 |
| | 2 | 0.42 | 0.44 | 0.05 | 0.12 |
| | 3 | 0.61 | 0.32 | 0.16 | 0.03 |
| | 5 | 0.4 | 0.54 | 0.14 | 0.08 |
| | 4 | 0.62 | 0.29 | 0.16 | 0.02 |
| | 7 | 0.74 | 0.23 | 0.08 | 0.04 |
| 4.0 | 1 | 0.73 | 0.28 | 0.04 | 0.02 |
| | 2 | 0.72 | 0.28 | 0.04 | 0.02 |
| | 3 | 0.68 | 0.29 | 0.04 | 0.02 |
| | 5 | 0.46 | 0.27 | 0.03 | 0.02 |
| | 6 | 0.38 | 0.3 | 0.01 | 0.01 |
| | 4 | 0.74 | 0.25 | 0.03 | 0.01 |
| | 7 | 0.68 | 0.2 | 0.03 | 0.01 |

**Base:** gpt-3.5, Prompt 7

| Model | Prompt | Common correct in base prompt | Common correct in comparing prompt | Common incorrect in base prompt | Common incorrect in comparing prompt |
|---|---|---|---|---|---|
| 3.5 | 1 | 0.43 | 0.79 | 0.06 | 0.14 |
| | 2 | 0.26 | 0.87 | 0.04 | 0.2 |
| | 3 | 0.42 | 0.69 | 0.1 | 0.04 |
| | 5 | 0.17 | 0.71 | 0.03 | 0.03 |
| | 6 | 0.23 | 0.74 | 0.04 | 0.08 |
| | 4 | 0.46 | 0.68 | 0.12 | 0.04 |
| 4.0 | 1 | 0.57 | 0.68 | 0.03 | 0.04 |
| | 2 | 0.56 | 0.69 | 0.03 | 0.04 |
| | 3 | 0.52 | 0.7 | 0.03 | 0.04 |
| | 5 | 0.36 | 0.67 | 0.02 | 0.03 |
| | 6 | 0.27 | 0.67 | 0.01 | 0.02 |
| | 4 | 0.58 | 0.63 | 0.04 | 0.03 |
| | 7 | 0.7 | 0.65 | 0.05 | 0.02 |

**Base:** gpt-4.0, Prompt 1

| Model | Prompt | Common correct in base prompt | Common correct in comparing prompt | Common incorrect in base prompt | Common incorrect in comparing prompt |
|---|---|---|---|---|---|
| 3.5 | 1 | 0.5 | 0.77 | 0.04 | 0.09 |
| | 2 | 0.3 | 0.84 | 0.03 | 0.13 |
| | 3 | 0.48 | 0.65 | 0.04 | 0.01 |
| | 5 | 0.18 | 0.64 | 0.02 | 0.02 |
| | 6 | 0.28 | 0.73 | 0.02 | 0.04 |
| | 4 | 0.51 | 0.63 | 0.05 | 0.01 |
| | 7 | 0.68 | 0.57 | 0.04 | 0.03 |
| 4.0 | 2 | 0.91 | 0.93 | 0.63 | 0.65 |
| | 3 | 0.75 | 0.84 | 0.33 | 0.33 |
| | 5 | 0.51 | 0.8 | 0.23 | 0.24 |
| | 6 | 0.37 | 0.76 | 0.18 | 0.28 |
| | 4 | 0.83 | 0.75 | 0.4 | 0.27 |
| | 7 | 0.74 | 0.57 | 0.13 | 0.05 |

**Base:** gpt-4.0, Prompt 2

| Model | Prompt | Common correct in base prompt | Common correct in comparing prompt | Common incorrect in base prompt | Common incorrect in comparing prompt |
|---|---|---|---|---|---|
| 3.5 | 1 | 0.52 | 0.77 | 0.05 | 0.09 |
| | 2 | 0.31 | 0.85 | 0.03 | 0.13 |
| | 3 | 0.48 | 0.65 | 0.05 | 0.01 |
| | 5 | 0.18 | 0.63 | 0.02 | 0.02 |
| | 6 | 0.28 | 0.72 | 0.02 | 0.04 |
| | 4 | 0.51 | 0.62 | 0.05 | 0.01 |
| | 7 | 0.69 | 0.56 | 0.04 | 0.03 |
| 4.0 | 1 | 0.93 | 0.91 | 0.65 | 0.63 |
| | 3 | 0.76 | 0.84 | 0.34 | 0.32 |
| | 5 | 0.5 | 0.77 | 0.23 | 0.24 |
| | 6 | 0.37 | 0.74 | 0.17 | 0.27 |
| | 4 | 0.84 | 0.74 | 0.37 | 0.24 |
| | 7 | 0.75 | 0.57 | 0.14 | 0.06 |

**Base:** gpt-4.0, Prompt 3

| Model | Prompt | Common correct in base prompt | Common correct in comparing prompt | Common incorrect in base prompt | Common incorrect in comparing prompt |
|---|---|---|---|---|---|
| 3.5 | 1 | 0.53 | 0.72 | 0.04 | 0.08 |
| | 2 | 0.33 | 0.82 | 0.03 | 0.13 |
| | 3 | 0.48 | 0.59 | 0.05 | 0.01 |
| | 5 | 0.18 | 0.58 | 0.02 | 0.02 |
| | 6 | 0.29 | 0.68 | 0.02 | 0.04 |
| | 4 | 0.51 | 0.56 | 0.06 | 0.02 |
| | 7 | 0.7 | 0.52 | 0.04 | 0.03 |
| 4.0 | 1 | 0.84 | 0.75 | 0.33 | 0.33 |
| | 2 | 0.84 | 0.76 | 0.32 | 0.34 |
| | 5 | 0.5 | 0.7 | 0.19 | 0.2 |
| | 6 | 0.38 | 0.69 | 0.13 | 0.21 |
| | 4 | 0.84 | 0.68 | 0.36 | 0.24 |
| | 7 | 0.76 | 0.52 | 0.12 | 0.05 |

**Base:** gpt-4.0, Prompt 4

| Model | Prompt | Common correct in base prompt | Common correct in comparing prompt | Common incorrect in base prompt | Common incorrect in comparing prompt |
|---|---|---|---|---|---|
| 3.5 | 1 | 0.46 | 0.77 | 0.03 | 0.1 |
| | 2 | 0.28 | 0.86 | 0.02 | 0.14 |
| | 3 | 0.43 | 0.65 | 0.04 | 0.02 |
| | 5 | 0.17 | 0.67 | 0.01 | 0.02 |
| | 6 | 0.25 | 0.74 | 0.01 | 0.03 |
| | 4 | 0.47 | 0.63 | 0.03 | 0.01 |
| | 7 | 0.63 | 0.58 | 0.03 | 0.04 |
| 4.0 | 1 | 0.75 | 0.83 | 0.27 | 0.4 |
| | 2 | 0.74 | 0.84 | 0.24 | 0.37 |
| | 3 | 0.68 | 0.84 | 0.24 | 0.36 |
| | 5 | 0.46 | 0.8 | 0.19 | 0.29 |
| | 6 | 0.34 | 0.76 | 0.13 | 0.32 |
| | 7 | 0.69 | 0.59 | 0.12 | 0.08 |

**Base:** gpt-4.0, Prompt 5

| Model | Prompt | Common correct in base prompt | Common correct in comparing prompt | Common incorrect in base prompt | Common incorrect in comparing prompt |
|---|---|---|---|---|---|
| 3.5 | 1 | 0.51 | 0.5 | 0.03 | 0.06 |
| | 2 | 0.33 | 0.59 | 0.02 | 0.1 |
| | 3 | 0.45 | 0.39 | 0.03 | 0.01 |
| | 5 | 0.18 | 0.42 | 0.01 | 0.01 |
| | 6 | 0.27 | 0.46 | 0.02 | 0.03 |
| | 4 | 0.52 | 0.41 | 0.02 | 0.01 |
| | 7 | 0.67 | 0.36 | 0.03 | 0.02 |
| 4.0 | 1 | 0.8 | 0.51 | 0.24 | 0.23 |
| | 2 | 0.77 | 0.5 | 0.24 | 0.23 |
| | 3 | 0.7 | 0.5 | 0.2 | 0.19 |
| | 6 | 0.49 | 0.64 | 0.23 | 0.34 |
| | 4 | 0.8 | 0.46 | 0.29 | 0.19 |
| | 7 | 0.74 | 0.36 | 0.08 | 0.03 |

**Base:** gpt-4.0, Prompt 6

| Model | Prompt | Common correct in base prompt | Common correct in comparing prompt | Common incorrect in base prompt | Common incorrect in comparing prompt |
|---|---|---|---|---|---|
| 3.5 | 1 | 0.52 | 0.38 | 0.03 | 0.04 |
| | 2 | 0.36 | 0.49 | 0.02 | 0.05 |
| | 3 | 0.48 | 0.32 | 0.03 | 0.01 |
| | 5 | 0.19 | 0.34 | 0.01 | 0 |
| | 6 | 0.3 | 0.38 | 0.01 | 0.01 |
| | 4 | 0.5 | 0.3 | 0.03 | 0.01 |
| | 7 | 0.67 | 0.27 | 0.02 | 0.01 |
| 4.0 | 1 | 0.76 | 0.37 | 0.28 | 0.18 |
| | 2 | 0.74 | 0.37 | 0.27 | 0.17 |
| | 3 | 0.69 | 0.38 | 0.21 | 0.13 |
| | 5 | 0.64 | 0.49 | 0.34 | 0.23 |
| | 4 | 0.76 | 0.34 | 0.32 | 0.13 |
| | 7 | 0.72 | 0.27 | 0.12 | 0.03 |

**Base:** gpt-4.0, Prompt 7

| Model | Prompt | Common correct in base prompt | Common correct in comparing prompt | Common incorrect in base prompt | Common incorrect in comparing prompt |
|---|---|---|---|---|---|
| 3.5 | 1 | 0.4 | 0.81 | 0.02 | 0.09 |
| | 2 | 0.24 | 0.87 | 0.01 | 0.13 |
| | 3 | 0.36 | 0.66 | 0.03 | 0.02 |
| | 5 | 0.15 | 0.68 | 0 | 0.01 |
| | 6 | 0.2 | 0.7 | 0.01 | 0.03 |
| | 4 | 0.41 | 0.67 | 0.04 | 0.03 |
| | 7 | 0.65 | 0.71 | 0.02 | 0.06 |
| 4.0 | 1 | 0.57 | 0.75 | 0.05 | 0.14 |
| | 2 | 0.57 | 0.77 | 0.06 | 0.15 |
| | 3 | 0.52 | 0.76 | 0.05 | 0.12 |
| | 5 | 0.36 | 0.75 | 0.03 | 0.09 |
| | 6 | 0.27 | 0.74 | 0.03 | 0.12 |
| | 4 | 0.59 | 0.7 | 0.08 | 0.13 |

# APPENDIX 3: Pairwise concordance across all models and prompts using the BIOC-GS corpus

**Base:** gpt-3.5, Prompt 1

| Model | Prompt | Common correct in base prompt | Common correct in comparing prompt | Common incorrect in base prompt | Common incorrect in comparing prompt |
|---|---|---|---|---|---|
| 3.5 | 2 | 0.42 | 0.97 | 0.14 | 0.65 |
| 3.5 | 3 | 0.71 | 0.3 | 0.28 | 0.04 |
| 3.5 | 5 | 0.45 | 0.32 | 0.21 | 0.08 |
| 3.5 | 6 | 0.55 | 0.34 | 0.17 | 0.05 |
| 3.5 | 4 | 0.71 | 0.27 | 0.3 | 0.04 |
| 3.5 | 7 | 0.82 | 0.18 | 0.15 | 0.04 |
| 4.0 | 1 | 0.81 | 0.23 | 0.07 | 0.02 |
| 4.0 | 2 | 0.8 | 0.22 | 0.1 | 0.02 |
| 4.0 | 3 | 0.79 | 0.23 | 0.1 | 0.02 |
| 4.0 | 5 | 0.55 | 0.26 | 0.05 | 0 |
| 4.0 | 6 | 0.56 | 0.25 | 0.06 | 0.02 |
| 4.0 | 4 | 0.87 | 0.2 | 0.11 | 0.02 |
| 4.0 | 7 | 0.42 | 0.22 | 0.06 | 0.01 |

**Base:** gpt-3.5, Prompt 2

| Model | Prompt | Common correct in base prompt | Common correct in comparing prompt | Common incorrect in base prompt | Common incorrect in comparing prompt |
|---|---|---|---|---|---|
| 3.5 | 1 | 0.97 | 0.42 | 0.65 | 0.14 |
| 3.5 | 3 | 0.81 | 0.15 | 0.24 | 0.01 |
| 3.5 | 5 | 0.56 | 0.17 | 0.29 | 0.02 |
| 3.5 | 6 | 0.58 | 0.16 | 0.24 | 0.01 |
| 3.5 | 4 | 0.75 | 0.12 | 0.18 | 0 |
| 3.5 | 7 | 0.81 | 0.08 | 0.06 | 0 |
| 4.0 | 1 | 0.83 | 0.1 | 0.06 | 0 |
| 4.0 | 2 | 0.83 | 0.1 | 0.06 | 0 |
| 4.0 | 3 | 0.83 | 0.1 | 0.12 | 0.01 |
| 4.0 | 5 | 0.56 | 0.11 | 0 | 0 |
| 4.0 | 6 | 0.61 | 0.12 | 0.06 | 0 |
| 4.0 | 4 | 0.89 | 0.09 | 0.06 | 0 |
| 4.0 | 7 | 0.85 | 0.24 | 0.13 | 0.01 |

**Base:** gpt-3.5, Prompt 3

| Model | Prompt | Common correct in base prompt | Common correct in comparing prompt | Common incorrect in base prompt | Common incorrect in comparing prompt |
|---|---|---|---|---|---|
| 3.5 | 1 | 0.3 | 0.71 | 0.04 | 0.28 |
| | 2 | 0.15 | 0.81 | 0.01 | 0.24 |
| | 5 | 0.37 | 0.63 | 0.07 | 0.17 |
| | 6 | 0.44 | 0.65 | 0.11 | 0.2 |
| | 4 | 0.72 | 0.63 | 0.32 | 0.26 |
| | 7 | 0.78 | 0.4 | 0.08 | 0.14 |
| 4.0 | 1 | 0.71 | 0.47 | 0.04 | 0.06 |
| | 2 | 0.71 | 0.47 | 0.05 | 0.08 |
| | 3 | 0.7 | 0.47 | 0.04 | 0.07 |
| | 5 | 0.45 | 0.5 | 0.02 | 0.01 |
| | 6 | 0.44 | 0.47 | 0.03 | 0.07 |
| | 4 | 0.81 | 0.45 | 0.05 | 0.05 |
| | 7 | 0.4 | 0.51 | 0.08 | 0.08 |

**Base:** gpt-3.5, Prompt 4

| Model | Prompt | Common correct in base prompt | Common correct in comparing prompt | Common incorrect in base prompt | Common incorrect in comparing prompt |
|---|---|---|---|---|---|
| 3.5 | 1 | 0.27 | 0.71 | 0.04 | 0.3 |
| | 2 | 0.12 | 0.75 | 0 | 0.18 |
| | 3 | 0.63 | 0.72 | 0.26 | 0.32 |
| | 5 | 0.36 | 0.69 | 0.07 | 0.2 |
| | 6 | 0.41 | 0.68 | 0.1 | 0.22 |
| | 7 | 0.8 | 0.47 | 0.06 | 0.13 |
| 4.0 | 1 | 0.78 | 0.58 | 0.05 | 0.08 |
| | 2 | 0.76 | 0.57 | 0.05 | 0.09 |
| | 3 | 0.76 | 0.58 | 0.03 | 0.06 |
| | 5 | 0.49 | 0.61 | 0.02 | 0.01 |
| | 6 | 0.49 | 0.59 | 0.02 | 0.07 |
| | 4 | 0.86 | 0.53 | 0.05 | 0.07 |
| | 7 | 0.41 | 0.58 | 0.05 | 0.07 |

**Base:** gpt-3.5, Prompt 5

| Model | Prompt | Common correct in base prompt | Common correct in comparing prompt | Common incorrect in base prompt | Common incorrect in comparing prompt |
|---|---|---|---|---|---|
| 3.5 | 1 | 0.32 | 0.45 | 0.08 | 0.21 |
|  | 2 | 0.17 | 0.56 | 0.02 | 0.29 |
|  | 3 | 0.63 | 0.37 | 0.17 | 0.07 |
|  | 6 | 0.55 | 0.48 | 0.26 | 0.19 |
|  | 4 | 0.69 | 0.36 | 0.2 | 0.07 |
|  | 7 | 0.78 | 0.24 | 0.11 | 0.08 |
| 4.0 | 1 | 0.72 | 0.28 | 0.06 | 0.04 |
|  | 2 | 0.74 | 0.29 | 0.09 | 0.05 |
|  | 3 | 0.71 | 0.29 | 0.09 | 0.06 |
|  | 5 | 0.45 | 0.3 | 0.05 | 0.01 |
|  | 6 | 0.43 | 0.27 | 0.05 | 0.04 |
|  | 4 | 0.85 | 0.28 | 0.09 | 0.04 |
|  | 7 | 0.33 | 0.25 | 0.06 | 0.03 |

**Base:** gpt-3.5, Prompt 6

| Model | Prompt | Common correct in base prompt | Common correct in comparing prompt | Common incorrect in base prompt | Common incorrect in comparing prompt |
|---|---|---|---|---|---|
| 3.5 | 1 | 0.34 | 0.55 | 0.05 | 0.17 |
|  | 2 | 0.16 | 0.58 | 0.01 | 0.24 |
|  | 3 | 0.65 | 0.44 | 0.2 | 0.11 |
|  | 5 | 0.48 | 0.55 | 0.19 | 0.26 |
|  | 4 | 0.68 | 0.41 | 0.22 | 0.1 |
|  | 7 | 0.74 | 0.26 | 0.05 | 0.05 |
| 4.0 | 1 | 0.75 | 0.33 | 0.05 | 0.04 |
|  | 2 | 0.76 | 0.34 | 0.05 | 0.04 |
|  | 3 | 0.75 | 0.35 | 0.03 | 0.03 |
|  | 5 | 0.47 | 0.35 | 0.03 | 0.01 |
|  | 6 | 0.44 | 0.32 | 0.04 | 0.05 |
|  | 4 | 0.85 | 0.32 | 0.06 | 0.04 |
|  | 7 | 0.35 | 0.3 | 0.03 | 0.02 |

**Base:** gpt-3.5, Prompt 7

| Model | Prompt | Common correct in base prompt | Common correct in comparing prompt | Common incorrect in base prompt | Common incorrect in comparing prompt |
|---|---|---|---|---|---|
| 3.5 | 1 | 0.18 | 0.82 | 0.04 | 0.15 |
| | 2 | 0.08 | 0.81 | 0 | 0.06 |
| | 3 | 0.4 | 0.78 | 0.14 | 0.08 |
| | 5 | 0.24 | 0.78 | 0.08 | 0.11 |
| | 6 | 0.26 | 0.74 | 0.05 | 0.05 |
| | 4 | 0.47 | 0.8 | 0.13 | 0.06 |
| 4.0 | 1 | 0.61 | 0.77 | 0.08 | 0.07 |
| | 2 | 0.59 | 0.76 | 0.1 | 0.08 |
| | 3 | 0.6 | 0.79 | 0.08 | 0.08 |
| | 5 | 0.35 | 0.76 | 0.04 | 0.01 |
| | 6 | 0.37 | 0.76 | 0.04 | 0.05 |
| | 4 | 0.68 | 0.73 | 0.08 | 0.05 |
| | 7 | 0.3 | 0.74 | 0.1 | 0.06 |

**Base:** gpt-4.0, Prompt 1

| Model | Prompt | Common correct in base prompt | Common correct in comparing prompt | Common incorrect in base prompt | Common incorrect in comparing prompt |
|---|---|---|---|---|---|
| 3.5 | 1 | 0.23 | 0.81 | 0.02 | 0.07 |
| | 2 | 0.1 | 0.83 | 0 | 0.06 |
| | 3 | 0.47 | 0.71 | 0.06 | 0.04 |
| | 5 | 0.28 | 0.72 | 0.04 | 0.06 |
| | 6 | 0.33 | 0.75 | 0.04 | 0.05 |
| | 4 | 0.58 | 0.78 | 0.08 | 0.05 |
| | 7 | 0.77 | 0.61 | 0.07 | 0.08 |
| 4.0 | 2 | 0.94 | 0.95 | 0.67 | 0.65 |
| | 3 | 0.84 | 0.87 | 0.32 | 0.35 |
| | 5 | 0.5 | 0.84 | 0.19 | 0.06 |
| | 6 | 0.48 | 0.78 | 0.11 | 0.17 |
| | 4 | 0.93 | 0.78 | 0.33 | 0.25 |
| | 7 | 0.4 | 0.77 | 0.07 | 0.05 |

**Base:** gpt-4.0, Prompt 2

| Model | Prompt | Common correct in base prompt | Common correct in comparing prompt | Common incorrect in base prompt | Common incorrect in comparing prompt |
|---|---|---|---|---|---|
| 3.5 | 1 | 0.22 | 0.8 | 0.02 | 0.1 |
| | 2 | 0.1 | 0.83 | 0 | 0.06 |
| | 3 | 0.47 | 0.71 | 0.08 | 0.05 |
| | 5 | 0.29 | 0.74 | 0.05 | 0.09 |
| | 6 | 0.34 | 0.76 | 0.04 | 0.05 |
| | 4 | 0.57 | 0.76 | 0.09 | 0.05 |
| | 7 | 0.76 | 0.59 | 0.08 | 0.1 |
| 4.0 | 1 | 0.95 | 0.94 | 0.65 | 0.67 |
| | 3 | 0.83 | 0.86 | 0.32 | 0.36 |
| | 5 | 0.49 | 0.82 | 0.16 | 0.06 |
| | 6 | 0.48 | 0.78 | 0.12 | 0.19 |
| | 4 | 0.93 | 0.77 | 0.33 | 0.26 |
| | 7 | 0.39 | 0.76 | 0.09 | 0.07 |

**Base:** gpt-4.0, Prompt 3

| Model | Prompt | Common correct in base prompt | Common correct in comparing prompt | Common incorrect in base prompt | Common incorrect in comparing prompt |
|---|---|---|---|---|---|
| 3.5 | 1 | 0.23 | 0.79 | 0.02 | 0.1 |
| | 2 | 0.1 | 0.83 | 0.01 | 0.12 |
| | 3 | 0.47 | 0.7 | 0.07 | 0.04 |
| | 5 | 0.29 | 0.71 | 0.06 | 0.09 |
| | 6 | 0.35 | 0.75 | 0.03 | 0.03 |
| | 4 | 0.58 | 0.76 | 0.06 | 0.03 |
| | 7 | 0.79 | 0.6 | 0.08 | 0.08 |
| 4.0 | 1 | 0.87 | 0.84 | 0.35 | 0.32 |
| | 2 | 0.86 | 0.83 | 0.36 | 0.32 |
| | 5 | 0.51 | 0.83 | 0.16 | 0.05 |
| | 6 | 0.49 | 0.77 | 0.11 | 0.16 |
| | 4 | 0.95 | 0.77 | 0.37 | 0.26 |
| | 7 | 0.41 | 0.77 | 0.1 | 0.07 |

**Base:** gpt-4.0, Prompt 4

| Model | Prompt | Common correct in base prompt | Common correct in comparing prompt | Common incorrect in base prompt | Common incorrect in comparing prompt |
|---|---|---|---|---|---|
| 3.5 | 1 | 0.2 | 0.87 | 0.02 | 0.11 |
| | 2 | 0.09 | 0.89 | 0 | 0.06 |
| | 3 | 0.45 | 0.81 | 0.05 | 0.05 |
| | 5 | 0.28 | 0.85 | 0.04 | 0.09 |
| | 6 | 0.32 | 0.85 | 0.04 | 0.06 |
| | 4 | 0.53 | 0.86 | 0.07 | 0.05 |
| | 7 | 0.73 | 0.68 | 0.05 | 0.08 |
| 4.0 | 1 | 0.78 | 0.93 | 0.25 | 0.33 |
| | 2 | 0.77 | 0.93 | 0.26 | 0.33 |
| | 3 | 0.77 | 0.95 | 0.26 | 0.37 |
| | 5 | 0.45 | 0.91 | 0.11 | 0.05 |
| | 6 | 0.46 | 0.89 | 0.1 | 0.2 |
| | 7 | 0.38 | 0.87 | 0.08 | 0.08 |

**Base:** gpt-4.0, Prompt 5

| Model | Prompt | Common correct in base prompt | Common correct in comparing prompt | Common incorrect in base prompt | Common incorrect in comparing prompt |
|---|---|---|---|---|---|
| 3.5 | 1 | 0.26 | 0.55 | 0 | 0.05 |
| | 2 | 0.11 | 0.56 | 0 | 0 |
| | 3 | 0.5 | 0.45 | 0.01 | 0.02 |
| | 5 | 0.3 | 0.45 | 0.01 | 0.05 |
| | 6 | 0.35 | 0.47 | 0.01 | 0.03 |
| | 4 | 0.61 | 0.49 | 0.01 | 0.02 |
| | 7 | 0.76 | 0.35 | 0.01 | 0.04 |
| 4.0 | 1 | 0.84 | 0.5 | 0.06 | 0.19 |
| | 2 | 0.82 | 0.49 | 0.06 | 0.16 |
| | 3 | 0.83 | 0.51 | 0.05 | 0.16 |
| | 6 | 0.63 | 0.6 | 0.04 | 0.16 |
| | 4 | 0.91 | 0.45 | 0.05 | 0.11 |
| | 7 | 0.41 | 0.47 | 0.01 | 0.03 |

**Base:** gpt-4.0, Prompt 6

| Model | Prompt | Common correct in base prompt | Common correct in comparing prompt | Common incorrect in base prompt | Common incorrect in comparing prompt |
|---|---|---|---|---|---|
| 3.5 | 1 | 0.25 | 0.56 | 0.02 | 0.06 |
| | 2 | 0.12 | 0.61 | 0 | 0.06 |
| | 3 | 0.47 | 0.44 | 0.07 | 0.03 |
| | 5 | 0.27 | 0.43 | 0.04 | 0.05 |
| | 6 | 0.32 | 0.44 | 0.05 | 0.04 |
| | 4 | 0.59 | 0.49 | 0.07 | 0.02 |
| | 7 | 0.76 | 0.37 | 0.05 | 0.04 |
| 4.0 | 1 | 0.78 | 0.48 | 0.17 | 0.11 |
| | 2 | 0.78 | 0.48 | 0.19 | 0.12 |
| | 3 | 0.77 | 0.49 | 0.16 | 0.11 |
| | 5 | 0.6 | 0.63 | 0.16 | 0.04 |
| | 4 | 0.89 | 0.46 | 0.2 | 0.1 |
| | 7 | 0.31 | 0.37 | 0.05 | 0.02 |

**Base:** gpt-4.0, Prompt 7

| Model | Prompt | Common correct in base prompt | Common correct in comparing prompt | Common incorrect in base prompt | Common incorrect in comparing prompt |
|---|---|---|---|---|---|
| 3.5 | 1 | 0.22 | 0.42 | 0.01 | 0.06 |
| | 2 | 0.12 | 0.5 | 0 | 0 |
| | 3 | 0.51 | 0.4 | 0.08 | 0.08 |
| | 5 | 0.25 | 0.33 | 0.03 | 0.06 |
| | 6 | 0.3 | 0.35 | 0.02 | 0.03 |
| | 4 | 0.58 | 0.41 | 0.07 | 0.05 |
| | 7 | 0.74 | 0.3 | 0.06 | 0.1 |
| 4.0 | 1 | 0.77 | 0.4 | 0.05 | 0.07 |
| | 2 | 0.76 | 0.39 | 0.07 | 0.09 |
| | 3 | 0.77 | 0.41 | 0.07 | 0.1 |
| | 5 | 0.47 | 0.41 | 0.03 | 0.01 |
| | 6 | 0.37 | 0.31 | 0.02 | 0.05 |
| | 4 | 0.87 | 0.38 | 0.08 | 0.08 |

# APPENDIX 4: Top 5 incorrectly extracted HPO IDs across all experiments

| Prompt | gpt-3.5 | gpt-4.0 |
|---|---|---|
| 1 | Decreased body weight (HP:0004325)<br>Intellectual disability, profound (HP:0002187)<br>Intellectual disability, severe (HP:0010864)<br>Short stature (HP:0004322)<br>Nephropathy (HP:0000112) | Poor wound healing (HP:0001058)<br>Cerebral hamartoma (HP:0009731)<br>Abnormality of thumb epiphysis (HP:0009599)<br>Palmar pits (HP:0010610)<br>Unilateral vestibular schwannoma (HP:0009590) |
| 2 | Decreased body weight (HP:0004325)<br>Intellectual disability, severe (HP:0010864)<br>Diabetes insipidus (HP:0000873)<br>Facial hyperostosis (HP:0005465)<br>Thin vermilion border (HP:0000233) | Poor wound healing (HP:0001058)<br>Palmar pits (HP:0010610)<br>Cerebral hamartoma (HP:0009731)<br>Unilateral vestibular schwannoma (HP:0009590)<br>Intellectual disability, severe (HP:0010864) |
| 3 | Intellectual disability, profound (HP:0002187)<br>Joint hypermobility (HP:0001382)<br>Microcephaly (HP:0000252)<br>Hearing impairment (HP:0000365)<br>Abnormality of the nose (HP:0000366) | Cerebral hamartoma (HP:0009731)<br>Poor wound healing (HP:0001058)<br>Neoplasm of the skin (HP:0008069)<br>Intellectual disability, severe (HP:0010864)<br>Skin erosion (HP:0200041) |
| 4 | Intellectual disability, profound (HP:0002187)<br>Joint hypermobility (HP:0001382)<br>Abnormality of the nervous system (HP:0000707)<br>Phenotypic abnormality (HP:0000118)<br>Absent earlobe (HP:0000387) | Poor wound healing (HP:0001058)<br>Cerebral hamartoma (HP:0009731)<br>Autosomal dominant inheritance with maternal imprinting (HP:0012275)<br>Neoplasm of the skin (HP:0008069)<br>Failure to thrive (HP:0001508 |
| 5 | Nephropathy (HP:0000112)<br>Intellectual disability, profound (HP:0002187)<br>Teratoma (HP:0009792)<br>Abnormality of the nervous system (HP:0000707)<br>Seizure (HP:0001250) | Poor wound healing (HP:0001058)<br>Cerebral hamartoma (HP:0009731)<br>Autosomal dominant inheritance with maternal imprinting (HP:0012275)<br>Abnormal cellular physiology (HP:0011017)<br>Occasional neurofibromas (HP:0009595) |
| 6 | Renal cyst (HP:0000107)<br>Intellectual disability, profound (HP:0002187)<br>Nephropathy (HP:0000112)<br>Abnormality of the nervous system (HP:0000707)<br>Absent earlobe (HP:0000387) | Poor wound healing (HP:0001058)<br>Autosomal dominant inheritance with maternal imprinting (HP:0012275)<br>Hypotension (HP:0002615)<br>Cerebral hamartoma (HP:0009731)<br>Cataplexy (HP:0002524) |
| 7 | Intellectual disability, profound (HP:0002187)<br>Abnormality of the nervous system (HP:0000707)<br>HALLUCINATION (HP:0000000)<br>Spinal neurofibroma (HP:0009735)<br>Abnormality of the lower limb (HP:0002814) | Neoplasm of the skin (HP:0008069)<br>Poor wound healing (HP:0001058)<br>Unilateral vestibular schwannoma (HP:0009590)<br>Intellectual disability, severe (HP:0010864)<br>Cerebral hamartoma (HP:0009731) |